\documentclass[11pt,letterpaper]{article}
\pdfoutput=1
\usepackage{emnlp2017}
\usepackage{times}
\usepackage{url}
\usepackage{latexsym}

\usepackage{mathtools}
\usepackage{calc}
\usepackage{amsmath, amsthm, amssymb, amsfonts}
\emnlpfinalcopy

\def\emnlppaperid{1005}

\title{Fast and Accurate Entity Recognition with Iterated Dilated Convolutions}

\author{Emma Strubell \qquad Patrick Verga \qquad David Belanger \qquad Andrew McCallum\\
  College of Information and Computer Sciences \\
  University of Massachusetts Amherst \\
  {\tt \{strubell, pat, belanger, mccallum\}@cs.umass.edu}
}

\date{}

\begin{document}
\maketitle
\begin{abstract}

Today when many practitioners run basic NLP on the entire web and large-volume traffic, faster methods are paramount to saving time and energy costs.
Recent advances in GPU hardware have led to the emergence of bi-directional LSTMs as a standard method for obtaining per-token vector representations serving as input to labeling tasks such as NER (often followed by prediction in a linear-chain CRF). 
Though expressive and accurate, these models fail to fully exploit GPU parallelism, limiting their computational efficiency.
This paper proposes a faster alternative to Bi-LSTMs for NER: Iterated Dilated Convolutional Neural Networks (ID-CNNs), which have better capacity than traditional CNNs for large context and structured prediction.  
Unlike LSTMs whose sequential processing on sentences of length $N$ requires $O(N)$ time even in the face of parallelism, ID-CNNs permit fixed-depth convolutions to run in parallel across entire documents.
We describe a distinct combination of network structure, parameter sharing and training procedures that enable dramatic 14-20x test-time speedups while retaining accuracy comparable to the Bi-LSTM-CRF. Moreover, ID-CNNs trained to aggregate context from the entire document are even more accurate while maintaining 8x faster test time speeds.

\end{abstract}

\section{Introduction}


In order to democratize large-scale NLP and information extraction while minimizing our environmental footprint, we require fast, resource-efficient methods for sequence tagging tasks such as part-of-speech tagging and named entity recognition (NER). Speed is not sufficient of course: they must also be expressive enough to tolerate the tremendous lexical variation in input data. 

The massively parallel computation facilitated by GPU hardware has led to a surge of successful neural network architectures for sequence labeling~\citep{ling2015finding,ma2016end,chiu2016named,lample2016neural}. While these models are expressive and accurate, they fail to fully exploit the parallelism opportunities of a GPU, and thus their speed is limited. Specifically, they employ either recurrent neural networks (RNNs) for feature extraction, or Viterbi inference in a structured output model, both of which require sequential computation across the length of the input.

Instead, parallelized runtime independent of the length of the sequence saves time and energy costs, maximizing GPU resource usage and minimizing the amount of time it takes to train and evaluate models. Convolutional neural networks (CNNs) provide exactly this property~\citep{kim2014convolutional,kalchbrenner2014convolutional}. Rather than composing representations incrementally over each token in a sequence, they apply filters in parallel across the entire sequence at once.  Their computational cost grows with the number of layers, but not the input size, up to the memory and threading limitations of the hardware. This provides, for example, audio generation models that can be trained in parallel~\citep{vandenoord2016wavenet}.


Despite the clear computational advantages of CNNs, RNNs have become the standard method for composing deep representations of text. This is because a token encoded by a bidirectional RNN will incorporate evidence from the entire input sequence, but the CNN's representation is limited by the effective input width\footnote{What we call \emph{effective input width} here is known as the \emph{receptive field} in the vision literature, drawing an analogy to the visual receptive field of a neuron in the retina.} of the network: the size of the input context which is observed, directly or indirectly, by the representation of a token at a given layer in the network. Specifically, in a network composed of a series of stacked convolutional layers of convolution width $w$, the number $r$ of context tokens incorporated into a token's representation at a given layer $l$, is given by $r=l(w-1) + 1$. The number of layers required to incorporate the entire input context grows linearly with the length of the sequence. To avoid this scaling, one could pool representations across the sequence, but this is not appropriate for sequence labeling, since it reduces the output resolution of the representation.



\begin{figure}
\includegraphics[scale=1.0]{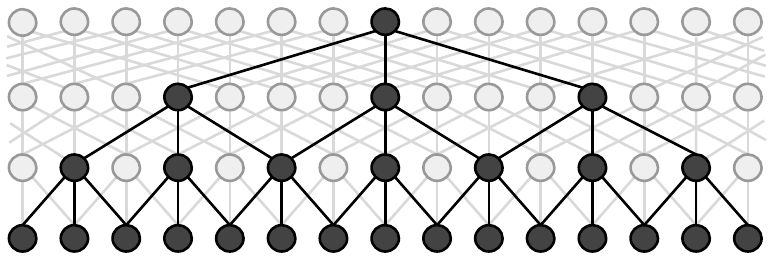}
\caption{A dilated CNN block with maximum dilation width 4 and filter width 3. Neurons contributing to a single highlighted neuron in the last layer are also highlighted. \label{dilated-block-fig}}
\end{figure}

In response, this paper presents an application of~\emph{dilated convolutions} \citep{yu2015multi} for sequence labeling (Figure \ref{dilated-block-fig}). For dilated convolutions, the effective input width can grow exponentially with the depth, with no loss in resolution at each layer and with a modest number of parameters to estimate. Like typical CNN layers, dilated convolutions operate on a sliding window of context over the sequence, but unlike conventional convolutions, the context need not be consecutive; the dilated window skips over every dilation width $d$ inputs. By stacking layers of dilated convolutions of exponentially increasing dilation width, we can expand the size of the effective input width to cover the entire length of most sequences using only a few layers: The size of the effective input width for a token at layer $l$ is now given by $2^{l+1}-1$. More concretely, just four stacked dilated convolutions of width 3 produces token representations with a n effective input width of 31 tokens -- longer than the average sentence length (23) in the Penn TreeBank. 

Our overall \textit{iterated dilated CNN} architecture (ID-CNN) repeatedly applies the same block of dilated convolutions to token-wise representations. This parameter sharing prevents overfitting and also provides opportunities to inject supervision on intermediate activations of the network. Similar to models that use logits produced by an RNN, the ID-CNN provides two methods for performing prediction: we can predict each token's label independently, or by running Viterbi inference in a chain structured graphical model. 

In experiments on CoNLL 2003 and OntoNotes 5.0 English NER, we demonstrate significant speed gains of our ID-CNNs over various recurrent models, while maintaining similar F1 performance. When performing prediction using independent classification, the ID-CNN consistently outperforms a bidirectional LSTM (Bi-LSTM), and performs on par with inference in a CRF with logits from a Bi-LSTM (Bi-LSTM-CRF). As an extractor of per-token logits for a CRF, our model out-performs the Bi-LSTM-CRF. We also apply ID-CNNs to entire documents, where independent token classification is as accurate as the Bi-LSTM-CRF while decoding almost 8$\times$ faster. The clear accuracy gains resulting from incorporating broader context suggest that these models could similarly benefit many other context-sensitive NLP tasks which have until now been limited by the computational complexity of existing context-rich models.\footnote{Our implementation in TensorFlow \citep{abadi2015tensorflow} is available at: \url{{https://github.com/iesl/dilated-cnn-ner}}}

\section{Background}
\subsection{Conditional Probability Models for Tagging}
\label{sec:prob-model}
Let $x = [x_1, \ldots, x_T]$ be our input text and $y = [y_1, \ldots, y_T]$ be per-token output tags. Let $D$ be the domain size of each $y_i$. We predict the most likely $y$, given a conditional model $P(y | x)$. 

This paper considers two factorizations of the conditional distribution. First, we have
\begin{equation}
P(y | x) = \prod_{t = 1}^T P(y_t | F(x)),
\label{eq:cond-ind}
\end{equation}
where the tags are conditionally independent given some features for x. Given these features, $O(D)$ prediction is simple and parallelizable across the length of the sequence. However, feature extraction may not necessarily be parallelizable. For example, RNN-based features require iterative passes along the length of $x$.

We also consider a linear-chain CRF model that couples all of $y$ together:
\begin{equation}
P(y | x) = \frac{1}{Z_x} \prod_{t = 1}^T \psi_t(y_t | F(x)) \psi_p(y_t,y_{t-1}),\label{eq:crf}
\end{equation}
where $\psi_{t}$ is a local factor, $\psi_p$ is a pairwise factor that scores consecutive tags, and $Z_x$ is the partition function~\citep{lafferty2001conditional}. To avoid overfitting, $\psi_p$ does not depend on the timestep $t$ or the input $x$ in our experiments. Prediction in this model requires global search using the $O(D^2 T)$ Viterbi algorithm. 

CRF prediction explicitly reasons about interactions among neighboring output tags, whereas prediction in the first model compiles this reasoning into the feature extraction step~\citep{liang2008structure}. The suitability of such compilation depends on the properties and quantity of the data. While CRF prediction requires non-trivial search in output space, it can guarantee that certain output constraints, such as for IOB tagging ~\citep{ramshaw1999text}, will always be satisfied. It may also have better sample complexity, as it imposes more prior knowledge about the structure of the interactions among the tags~\citep{JMLR:v17:15-501}. However, it has worse computational complexity than independent prediction.

\section{Dilated Convolutions}


CNNs in NLP are typically one-dimensional, applied to a sequence of vectors representing tokens rather than to a two-dimensional grid of vectors representing pixels. In this setting, a convolutional neural network layer is equivalent to applying an affine transformation, $W_c$ to a sliding window of width $r$ tokens on either side of each token in the sequence. Here, and throughout the paper, we do not explicitly write the bias terms in affine transformations. The convolutional operator applied to each token $x_t$ with output $c_t$ is defined as:
\begin{align}
c_t = W_c\bigoplus_{k=0}^{r}x_{t \pm k},
\end{align}
where $\oplus$ is vector concatenation.

Dilated convolutions perform the same operation, except rather than transforming adjacent inputs, the convolution is defined over a wider effective input width by skipping over $\delta$ inputs at a time, where $\delta$ is the dilation width. We define the dilated convolution operator:
\begin{align}
c_t = W_c\bigoplus_{k=0}^{r}x_{t \pm k\delta}.
\end{align}
A dilated convolution of width 1 is equivalent to a simple convolution. Using the same number of parameters as a simple convolution with the same radius (i.e. $W_c$ has the same dimensionality), the $\delta > 1$ dilated convolution incorporates broader context into the representation of a token than a simple convolution. 


\subsection{Multi-Scale Context Aggregation}


We can leverage the ability of dilated convolutions to incorporate global context without losing important local information by stacking dilated convolutions of increasing width. First described for pixel classification in computer vision, \citet{yu2015multi} achieve state-of-the-art results on image segmentation benchmarks by stacking dilated convolutions with exponentially increasing rates of dilation, a technique they refer to as \emph{multi-scale context aggregation}. By feeding the outputs of each dilated convolution as the input to the next, increasingly non-local information is incorporated into each pixel's representation. Performing a dilation-1 convolution in the first layer ensures that no pixels within the effective input width of any pixel are excluded. By doubling the dilation width at each layer, the size of the effective input width grows exponentially while the number of parameters grows only linearly with the number of layers, so a pixel representation quickly incorporates rich global evidence from the entire image. 

\section{Iterated Dilated CNNs}
\label{blocks-section}
Stacked dilated CNNs can easily incorporate global information from a whole sentence or document. For example, with a radius of 1 and 4 layers of dilated convolutions, the effective input width of each token is width 31, which exceeds the average sentence length (23) in the Penn TreeBank corpus. With a radius of size 2 and 8 layers of dilated convolutions, the effective input width exceeds 1,000 tokens, long enough to encode a full newswire document.

Unfortunately, simply increasing the depth of stacked dilated CNNs causes considerable over-fitting in our experiments. In response, we present Iterated Dilated CNNs (ID-CNNs), which instead apply the same small stack of dilated convolutions multiple times, each iterate taking as input the result of the last application. Repeatedly employing the same parameters in a recurrent fashion provides both broad effective input width and desirable generalization capabilities. We also obtain significant accuracy gains with a training objective that strives for accurate labeling after each iterate, allowing follow-on iterations to observe and resolve dependency violations.


\subsection{Model Architecture\label{model-section}}
The network takes as input a sequence of $T$ vectors ${\bf x_t}$, and outputs a sequence of per-class scores ${\bf h_t}$, which serve either as the local conditional distributions of Eqn.~\eqref{eq:cond-ind} or the local factors $\psi_t$ of Eqn.~\eqref{eq:crf}. 



We denote the $j$th dilated convolutional layer of dilation width $\delta$ as $D_\delta^{(j)}$. The first layer in the network is a dilation-1 convolution $D_1^{(0)}$ that transforms the input to a representation ${\bf i_t}$:
\begin{align}
{\bf i_t} = D_1^{(0)}{\bf x_t}
\end{align}
Next, $L_c$ layers of dilated convolutions of exponentially increasing dilation width are applied to ${\bf i_t}$, folding in increasingly broader context into the embedded representation of ${\bf x_t}$ at each layer. Let $r()$ denote the ReLU activation function \citep{glorot2011deep}. Beginning with ${\bf c_t}^{(0)} = {\bf i_t}$ we define the stack of layers with the following recurrence:
\begin{align}
{\bf c_t}^{(j)} &= r\left(D_{2^{L_c-1}}^{(j-1)}{\bf c_t}^{(j-1)}\right)
\end{align}
and add a final dilation-1 layer to the stack:
\begin{align}
{\bf c_t}^{(L_c+1)} &= r\left(D_1^{(L_c)}{\bf c_t}^{(L_c)}\right)
\end{align}
We refer to this stack of dilated convolutions as a \emph{block} $B(\cdot)$, which has output resolution equal to its input resolution. To incorporate even broader context without over-fitting, we avoid making $B$ deeper, and instead iteratively apply $B$ $L_b$ times, introducing no extra parameters. Starting with ${\bf b_t}^{(1)} = B\left({\bf i_t}\right)$:
\begin{align}
{\bf b_t}^{(k)} &= B\left({\bf b_t}^{(k-1)} \right)
\label{block-eqn}
\end{align}
We apply a simple affine transformation $W_o$ to this final representation to obtain per-class scores for each token ${\bf x_t}$:
\begin{align}
{\bf h_t}^{(L_b)} = W_o{\bf b_t}^{(L_b)}
\label{outputs-eqn}
\end{align}


\subsection{Training}


Our main focus is to apply the ID-CNN an encoder to produce per-token logits for the first conditional model described in Sec.~\ref{sec:prob-model}, where tags are conditionally independent given deep features, since this will enable prediction that is parallelizable across the length of the input sequence. Here, maximum likelihood training is straightforward because the likelihood decouples into the sum of the likelihoods of independent logistic regression problems for every tag, with natural parameters given by Eqn.~\eqref{outputs-eqn}:
\begin{equation}
\frac{1}{T}\sum_{t=1}^T \log P(y_t \mid {\bf h_t}^{(L_b)}) \label{eq:loss}
\end{equation}

We can also use the ID-CNN as logits for the CRF model (Eqn.~\eqref{eq:crf}), where the partition function and its gradient are computed using the forward-backward algorithm. 

We next present an alternative training method that helps bridge the gap between these two techniques. Sec.~\ref{sec:prob-model} identifies that the CRF has preferable sample complexity and accuracy since prediction directly reasons in the space of structured outputs. In response, we compile some of this reasoning in output space into ID-CNN feature extraction. Instead of explicit reasoning over output labels during inference, we train the network such that each block is predictive of output labels. Subsequent blocks learn to correct dependency violations of their predecessors, refining the final sequence prediction.

To do so, we first define predictions of the model after each of the $L_b$ applications of the block. Let ${\bf h_t}^{(k)}$ be the result of applying the matrix $W_o$ from~\eqref{outputs-eqn} to ${\bf b_t}^{(k)}$, the output of block $k$. We minimize the average of the losses for each application of the block: 
\begin{align}
\frac{1}{L_b}\sum_{k=1}^{L_b} \frac{1}{T}\sum_{t=1}^T \log P(y_t \mid {\bf h_t}^{(k)}). \label{eq:avg-loss}
\end{align}

By rewarding accurate predictions after each application of the block, we learn a model where later blocks are used to refine initial predictions. The loss also helps reduce the vanishing gradient problem~\citep{hochreiter1998vanishing} for deep architectures. Such an approach has been applied in a variety of contexts for training very deep networks in computer vision~\citep{romero2014fitnets,szegedy2015going,lee2015deeply,gulccehre2016knowledge}, but not to our knowledge in NLP.

We apply dropout~\citep{srivastava2014dropout} to the raw inputs ${\bf x_t}$ and to each block's output ${\bf b_t}^{(b)}$ to help prevent overfitting. The version of dropout typically used in practice has the undesirable property that the randomized predictor used at train time differs from the fixed one used at test time. \citet{ma2017dropout} present \textit{dropout with expectation-linear regularization}, which explicitly regularizes these two predictors to behave similarly. All of our best reported results include such regularization. This is the first investigation of the technique's effectiveness for NLP, including for RNNs. We encourage its further application.

\section{Related work}


The state-of-the art models for sequence labeling include an inference step that searches the space of possible output sequences of a chain-structured graphical model, or approximates this search with a beam \citep{collobert2011natural, weiss2015structured, lample2016neural, ma2016end, chiu2016named}. These outperform similar systems that use the same features, but independent local predictions. On the other hand, the greedy \emph{sequential prediction} \citep{daume2009search} approach of~\citet{ratinov2009design}, which employs lexicalized features, gazetteers, and word clusters, outperforms CRFs with similar features.



LSTMs \citep{hochreiter1997long} were used for NER as early as the CoNLL shared task in 2003 \citep{hammerton2003named, tjong2003introduction}. More recently, a wide variety of neural network architectures for NER have been proposed. \citet{collobert2011natural} employ a one-layer CNN with pre-trained word embeddings, capitalization and lexicon features, and CRF-based prediction. \citet{huang2015bidirectional} achieved state-of-the-art accuracy on part-of-speech, chunking and NER using a Bi-LSTM-CRF. \citet{lample2016neural} proposed two models which incorporated Bi-LSTM-composed character embeddings alongside words: a Bi-LSTM-CRF, and a greedy stack LSTM which uses a simple shift-reduce grammar to compose words into labeled entities. Their Bi-LSTM-CRF obtained the state-of-the-art on four languages without word shape or lexicon features. \citet{ma2016end} use CNNs rather than LSTMs to compose characters in a Bi-LSTM-CRF, achieving state-of-the-art performance on part-of-speech tagging and CoNLL NER without lexicons. \citet{chiu2016named} evaluate a similar network but propose a novel method for encoding lexicon matches, presenting results on CoNLL and OntoNotes NER. \citet{yang2016multi} use GRU-CRFs with GRU-composed character embeddings of words to train a single network on many tasks and languages. 

In general, distributed representations for text can provide useful generalization capabilities for NER systems, since they can leverage unsupervised pre-training of distributed word representations \citep{turian2010word,collobert2011natural,passos2014lexicon}. Though our models would also likely benefit from additional features such as character representations and lexicons, we focus on simpler models which use word-embeddings alone, leaving more elaborate input representations to future work. 

In these NER approaches, CNNs were used for low-level feature extraction that feeds into alternative architectures. Overall, end-to-end CNNs have mainly been used in NLP for sentence classification, where the output representation is lower resolution than that of the input~\citet{kim2014convolutional,kalchbrenner2014convolutional,zhang2015character,toutanova2015representing}. \citet{lei2015molding} present a CNN variant where convolutions adaptively skip neighboring words. While the flexibility of this model is powerful, its adaptive behavior is not well-suited to GPU acceleration. 

Our work draws on the use of dilated convolutions for image segmentation in the computer vision community \citep{yu2015multi,chen2015semantic}. Similar to our block, \citet{yu2015multi} employ a 
\textit{context-module} of stacked dilated convolutions of exponentially increasing dilation width. Dilated convolutions were recently applied to the task of speech generation \citep{vandenoord2016wavenet}, and concurrent with this work, \citet{kalchbrenner2016neural} posted a pre-print describing the similar ByteNet network for machine translation that uses dilated convolutions in the encoder and decoder components. Our basic model architecture is similar to that of the ByteNet encoder, except that the inputs to our model are tokens and not bytes. Additionally, we present a novel loss and parameter sharing scheme to facilitate training models on much smaller datasets than those used by \citet{kalchbrenner2016neural}. We are the first to use dilated convolutions for sequence labeling. 




The broad effective input width of the ID-CNN helps aggregate document-level context. \citet{ratinov2009design} incorporate document context in their greedy model by adding features based on tagged entities within a large, fixed window of tokens.  Prior work has also posed a structured model that couples predictions across the whole document~\citep{bunescu2004collective,sutton2004collective,finkel2005incorporating}. 







\section{Experimental Results}

We describe experiments on two benchmark English named entity recognition datasets. On CoNLL-2003 English NER, our ID-CNN performs on par with a Bi-LSTM not only when used to produce per-token logits for structured inference, but the ID-CNN with greedy decoding also performs on-par with the Bi-LSTM-CRF while running at more than 14 times the speed. We also observe a performance boost in almost all models when broadening the context to incorporate entire documents, achieving an average F1 of 90.65 on CoNLL-2003, out-performing the sentence-level model while still decoding at nearly 8 times the speed of the Bi-LSTM-CRF.

\subsection{Data and Evaluation}
We evaluate using labeled data from the CoNLL-2003 shared task \citep{tjong2003introduction} and OntoNotes 5.0 \citep{hovy2006ontonotes,pradhan2013towards}. Following previous work, we use the same OntoNotes data split used for co-reference resolution in the CoNLL-2012 shared task \citep{pradhan2012conll}.
For both datasets, we convert the IOB boundary encoding to BILOU as previous work found this encoding to result in improved performance \citep{ratinov2009design}. As in previous work we evaluate the performance of our models using segment-level micro-averaged F1 score. Hyperparameters that resulted in the best performance on the validation set were selected via grid search. A more detailed description of the data, evaluation, optimization and data pre-processing can be found in the Appendix.

\subsection{Baselines \label{baselines-sec}}

We compare our {\bf ID-CNN} against strong LSTM and CNN baselines: a {\bf Bi-LSTM} with local decoding, and one with CRF decoding ({\bf Bi-LSTM-CRF}). We also compare against a non-dilated CNN architecture with the same number of convolutional layers as our dilated network ({\bf 4-layer CNN}) and one with enough layers to incorporate an effective input width of the same size as that of the dilated network ({\bf 5-layer CNN}) to demonstrate that the dilated convolutions more effectively aggregate contextual information than simple convolutions (i.e. using fewer parameters). We also compare our document-level ID-CNNs to a baseline which does not share parameters between blocks ({\bf noshare}) and one that computes loss only at the last block, rather than after every iterated block of dilated convolutions ({\bf 1-loss}).

We do not compare with deeper or more elaborate CNN architectures for a number of reasons: 1) Fast train and test performance are highly desirable for NLP practitioners, and deeper models require more computation time 2) more complicated models tend to over-fit on this relatively small dataset and 3) most accurate deep CNN architectures repeatedly up-sample and down-sample the inputs. We do not compare to stacked LSTMs for similar reasons --- a single LSTM is already slower than a 4-layer CNN. Since our task is sequence labeling, we desire a model that maintains the token-level resolution of the input, making dilated convolutions an elegant solution.

\subsection{CoNLL-2003 English NER}
\subsubsection{Sentence-level prediction}

\begin{table}
\begin{tabular}{ll}
    Model & F1 \\ \hline \hline
    \citet{ratinov2009design} & 86.82 \\ 
    \citet{collobert2011natural} & 86.96 \\ 
    \citet{lample2016neural} & 90.33 \\ \hline 
    Bi-LSTM & 89.34 $\pm$ 0.28 \\
    4-layer CNN & 89.97 $\pm$ 0.20 \\
    5-layer CNN & 90.23 $\pm$ 0.16 \\
    ID-CNN & 90.32 $\pm$ 0.26 \\ \hline\hline
    \citet{collobert2011natural} & 88.67 \\
    \citet{passos2014lexicon} & 90.05 \\
    \citet{lample2016neural} & 90.20 \\ \hline 
    Bi-LSTM-CRF (re-impl) & 90.43 $\pm$ 0.12 \\
    ID-CNN-CRF & {\bf 90.54 $\pm$ 0.18} \\
  \end{tabular}

  \caption{F1 score of models observing sentence-level context. No models use character embeddings or lexicons. Top models are greedy, bottom models use Viterbi inference \label{sentence-summary}.}
  
\end{table}
Table \ref{sentence-summary} lists F1 scores of models predicting with sentence-level context on CoNLL-2003. For models that we trained, we report F1 and standard deviation obtained by averaging over 10 random restarts. The Viterbi-decoding Bi-LSTM-CRF and ID-CNN-CRF and greedy ID-CNN obtain the highest average scores, with the ID-CNN-CRF outperforming the Bi-LSTM-CRF by $0.11$ points of F1 on average, and the Bi-LSTM-CRF out-performing the greedy ID-CNN by $0.11$ as well. Our greedy ID-CNN outperforms the Bi-LSTM and the 4-layer CNN, which uses the same number of parameters as the ID-CNN, and performs similarly to the 5-layer CNN which uses more parameters but covers the same effective input width. All CNN models out-perform the Bi-LSTM when paired with greedy decoding, suggesting that CNNs are better token encoders than Bi-LSTMs for independent logistic regression. When paired with Viterbi decoding, our ID-CNN performs on par with the Bi-LSTM, showing that the ID-CNN is also an effective token encoder for structured inference.

Our ID-CNN is not only a better token encoder than the Bi-LSTM but it is also faster. Table \ref{sentence-speed-table} lists relative decoding times on the CoNLL development set, compared to the Bi-LSTM-CRF. We report decoding times using the fastest batch size for each method.\footnote{For each model, we tried batch sizes $b = 2^i$ with $i = 0 ... 11$. At scale, speed should increase with batch size, as we could compose each batch of as many sentences of the same length as would fit in GPU memory, requiring no padding and giving CNNs and ID-CNNs even more of a speed advantage.}

The ID-CNN model decodes nearly 50\% faster than the Bi-LSTM. With Viterbi decoding, the gap closes somewhat but the ID-CNN-CRF still comes out ahead, about 30\% faster than the Bi-LSTM-CRF. The most vast speed improvements come when comparing the greedy ID-CNN to the Bi-LSTM-CRF -- our ID-CNN is more than 14 times faster than the Bi-LSTM-CRF at test time, with comparable accuracy. The 5-layer CNN, which observes the same effective input width as the ID-CNN but with more parameters, performs at about the same speed as the ID-CNN in our experiments. With a better implementation of dilated convolutions than currently included in TensorFlow, we would expect the ID-CNN to be notably faster than the 5-layer CNN.




\begin{table}
\begin{center}
\begin{tabular}{ll}
    Model & Speed \\ \hline \hline
    Bi-LSTM-CRF & 1$\times$ \\ 
    Bi-LSTM & 9.92$\times$ \\ 
    ID-CNN-CRF & 1.28$\times$ \\ 
    5-layer CNN & 12.38$\times$ \\ 
    ID-CNN & 14.10$\times$ \\ 
  \end{tabular}
  \end{center}
  \caption{Relative test-time speed of sentence models, using the fastest batch size for each model.\footnotemark
  \label{sentence-speed-table}}
\end{table}
\footnotetext{Our ID-CNN could see up to 18$\times$ speed-up with a less naive implementation than is included in TensorFlow as of this writing.}

We emphasize the importance of the dropout regularizer of \citet{ma2017dropout} in Table \ref{dr-table}, where we observe increased F1 for every model trained with expectation-linear dropout regularization. Dropout is important for training neural network models that generalize well, especially on relatively small NLP datasets such as CoNLL-2003. We recommend this regularizer as a simple and helpful tool for practitioners training neural networks for NLP.

\begin{table}
\begin{tabular}{lll}
    Model & w/o DR & w/ DR \\ \hline \hline
    Bi-LSTM & 88.89 $\pm$ 0.30 & {\bf 89.34} $\pm$ 0.28 \\
    4-layer CNN & 89.74 $\pm$ 0.23 & {\bf 89.97} $\pm$ 0.20\\
    5-layer CNN & 89.93 $\pm$ 0.32 & {\bf 90.23} $\pm$ 0.16\\
    Bi-LSTM-CRF & 90.01 $\pm$ 0.23 & {\bf 90.43} $\pm$ 0.12 \\
    4-layer ID-CNN & 89.65 $\pm$ 0.30 & {\bf 90.32} $\pm$ 0.26 \\
  \end{tabular}
  \caption{Comparison of models trained with and without expectation-linear dropout regularization (DR). DR improves all models. \label{dr-table}}
\end{table}

\subsubsection{Document-level prediction}

In Table \ref{doc-summary-table} we show that adding document-level context improves every model on CoNLL-2003. Incorporating document-level context further improves our greedy ID-CNN model, attaining 90.65 average F1. We believe this model sees greater improvement with the addition of document-level context than the Bi-LSTM-CRF due to the ID-CNN learning a feature function better suited for representing broad context, in contrast with the Bi-LSTM which, though better than a simple RNN at encoding long memories of sequences, may reach its limit when provided with sequences more than 1,000 tokens long such as entire documents.

\begin{table}
\begin{tabular}{ll}
    Model & F1 \\ \hline \hline
    4-layer ID-CNN (sent) & 90.32 $\pm$ 0.26 \\ 
    Bi-LSTM-CRF (sent) & 90.43 $\pm$ 0.12 \\ \hline 
    4-layer CNN $\times$ 3 & 90.32 $\pm$ 0.32 \\ 
    5-layer CNN $\times$ 3 & 90.45 $\pm$ 0.21 \\ 
    Bi-LSTM & 89.09 $\pm$ 0.19 \\ 
    Bi-LSTM-CRF & 90.60 $\pm$ 0.19 \\ 
    ID-CNN & {\bf 90.65 $\pm$ 0.15} \\ 
  \end{tabular}
  \caption{F1 score of models trained to predict document-at-a-time. Our greedy ID-CNN model performs as well as the Bi-LSTM-CRF.\label{doc-summary-table}}
\end{table}

We also note that our combination of training objective (Eqn. \ref{eq:avg-loss}) and tied parameters (Eqn. \ref{block-eqn}) more effectively learns to aggregate this broad context than a vanilla cross-entropy loss or deep CNN back-propagated from the final neural network layer. Table \ref{doc-loss-table} compares models trained to incorporate entire document context using the document baselines described in Section \ref{baselines-sec}.

\begin{table}
\begin{center}
\scalebox{0.92}{
\begin{tabular}{lll}
    Model & F1 \\ \hline \hline
    ID-CNN noshare & 89.81 $\pm$ 0.19 \\
    ID-CNN 1-loss & 90.06 $\pm$ 0.19 \\
    ID-CNN & {\bf 90.65 $\pm$ 0.15} \\
  \end{tabular}
  }
\end{center}
  \caption{Comparing ID-CNNs with 1) back-propagating loss only from the final layer ({\bf 1-loss}) and 2) untied parameters across blocks ({\bf noshare})}
  \label{doc-loss-table}
\end{table}

In Table \ref{doc-speed-table} we show that, in addition to being more accurate, our ID-CNN model is also much faster than the Bi-LSTM-CRF when incorporating context from entire documents, decoding at almost 8 times the speed. On these long sequences, it also tags at more than 4.5 times the speed of the greedy Bi-LSTM, demonstrative of the benefit of our ID-CNNs context-aggregating computation that does not depend on the length of the sequence.


\begin{table}
\begin{center}
\scalebox{0.92}{
\begin{tabular}{ll}
    Model & Speed\\ \hline \hline
    Bi-LSTM-CRF & 1$\times$ \\ 
    Bi-LSTM & 4.60$\times$ \\ 
    ID-CNN & 7.96$\times$ \\ 
  \end{tabular}
  }
    \end{center}
  \caption{Relative test-time speed of document models (fastest batch size for each model).}
  \label{doc-speed-table}
\end{table}

\subsection{OntoNotes 5.0 English NER}

We observe similar patterns on OntoNotes as we do on CoNLL. Table \ref{summary-table-onto} lists overall F1 scores of our models compared to those in the existing literature. The greedy Bi-LSTM out-performs the lexicalized greedy model of \citet{ratinov2009design}, and our ID-CNN out-performs the Bi-LSTM as well as the more complex model of \citet{durrett2014joint} which leverages the parallel co-reference annotation available in the OntoNotes corpus to predict named entities jointly with entity linking and co-reference. Our greedy model is out-performed by the Bi-LSTM-CRF reported in \citet{chiu2016named} as well as our own re-implementation, which appears to be the new state-of-the-art on this dataset. 

\begin{table}
\scalebox{0.92}{
\begin{tabular}{lll}
    Model & F1 & Speed\\ \hline \hline
    \citet{ratinov2009design}\footnotemark & 83.45 \\
    \citet{durrett2014joint} & 84.04 \\
    \citet{chiu2016named} & 86.19 $\pm$ 0.25 \\ \hline 
    Bi-LSTM-CRF & 86.99 $\pm$ 0.22 & 1$\times$ \\ 
    Bi-LSTM-CRF-Doc & 86.81 $\pm$ 0.18  & 1.32$\times$\\ 
    Bi-LSTM & 83.76 $\pm$ 0.10 & 24.44$\times$ \\ \hline 
    ID-CNN-CRF (1 block) & 86.84 $\pm$ 0.19 & 1.83$\times$ \\ 
    ID-CNN-Doc (3 blocks) & 85.76 $\pm$ 0.13  & 21.19$\times$ \\ 
    ID-CNN (3 blocks) & 85.27 $\pm$ 0.24 & 13.21$\times$ \\ 
    ID-CNN (1 block) & 84.28 $\pm$ 0.10 & 26.01$\times$ \\ 
  \end{tabular}
  }
  \caption{F1 score of sentence and document models on OntoNotes.\label{summary-table-onto}}
\end{table}
\footnotetext{\protect{Results as reported in \citet{durrett2014joint} as this data split
    did not exist at the time of publication.}}

The gap between our greedy model and those using Viterbi decoding is wider than on CoNLL. We believe this is due to the more diverse set of entities in OntoNotes, which also tend to be much longer -- the average length of a multi-token named entity segment in CoNLL is about one token shorter than in OntoNotes. These long entities benefit more from explicit structured constraints enforced in Viterbi decoding. Still, our ID-CNN outperforms all other greedy methods, achieving our goal of learning a better token encoder for structured prediction.

Incorporating greater context significantly boosts the score of our greedy model on OntoNotes, whereas the Bi-LSTM-CRF performs more poorly. In Table \ref{summary-table-onto}, we also list the F1 of our ID-CNN model and the Bi-LSTM-CRF model trained on entire document context. For the first time, we see the score decrease when more context is added to the Bi-LSTM-CRF model, though the ID-CNN, whose sentence model a lower score than that of the Bi-LSTM-CRF, sees an increase. We believe the decrease in the Bi-LSTM-CRF model occurs because of the nature of the OntoNotes dataset compared to CoNLL-2003: CoNLL-2003 contains a particularly high proportion of ambiguous entities,\footnote{According to the ACL Wiki page on CoNLL-2003: ``The corpus contains a very high ratio of metonymic references (city names standing for sport teams)''} perhaps leading to more benefit from document context that helps with disambiguation. In this scenario, adding the wider context may just add noise to the high-scoring Bi-LSTM-CRF model, whereas the less accurate dilated model can still benefit from the refined predictions of the iterated dilated convolutions.

\section{Conclusion}

We present iterated dilated convolutional neural networks, fast token encoders that efficiently aggregate broad context without losing resolution. These provide impressive speed improvements for sequence labeling, particularly when processing entire documents at a time. In the future we hope to extend this work to NLP tasks with richer structured output, such as parsing.

\section*{Acknowledgments}
We thank Subhransu Maji and Luke Vilnis for helpful discussions, and Brendan O'Connor, Yoav Goldberg, the UMass NLP reading group and many anonymous reviewers for constructive comments on various drafts of the paper. We are also grateful to Guillaume Lample for sharing his pre-trained word embeddings. This work was supported in part by the Center for Intelligent Information Retrieval, in part by DARPA under agreement number FA8750-13-2-0020, in part by Defense Advanced Research Agency (DARPA) contract number HR0011-15-2-0036, in part by the National Science Foundation (NSF) grant number DMR-1534431, and in part by the National Science Foundation (NSF) grant number IIS-1514053. The U.S. Government is authorized to reproduce and distribute reprints for Governmental purposes notwithstanding any copyright notation thereon. Any opinions, findings and conclusions or recommendations expressed in this material are those of the authors and do not necessarily reflect those of the sponsor.

\bibliography{emnlp2017}
\bibliographystyle{emnlp_natbib}

\clearpage
\newpage

\appendix

\section{Appendix}

\subsection{Optimization and data pre-processing}
Our models are trained end-to-end using backpropagation and mini-batched Adam \citep{kingma2014adam} SGD. We use dropout regularization \citep{srivastava2014dropout} on the input embeddings and final dilation layer of each block, along with the dropout regularizer described in \citet{ma2017dropout} using a single Monte Carlo sample for each training example. We also found word dropout \citep{dai2015semi,lample2016neural} crucial for learning a high-quality representation for out-of-vocabulary words. We used the modified version of identity initialization \citep{le2015simple} reported by \citet{yu2015multi} to initialize our dilated layers, which we found to perform the best in initial experiments compared to orthogonal and Xavier initialization \citep{glorot2010understanding}. Since our models use the same number of filters in each dilated layer, this initialization simplifies to setting the parameters corresponding to the central token to the identity matrix, and all other parameters (corresponding to left and right context) to zero. All other layers (embeddings, projections) were initialized using normally distributed Xavier initialization. 

As in previous work, we found that initializing the word embedding lookup table with pre-trained embeddings was vital to achieve good performance. In initial experiments, we found the 100-dimensional skip-n-gram \citep{wang2015not} embeddings of \citet{lample2016neural} to outperform the 50-dimensional word embeddings of \citet{collobert2011natural}, and so we use these 100-dimensional embeddings in all experiments. We concatenate a 5-dimensional word shape vector based on whether the token was all capitalized, not capitalized, first-letter capitalized or contained a capital letter. We preprocessed the data by replacing all digits with 0, but did not lowercase thus our embeddings are case-sensitive.

We use the parameters of the trained sentence models to initialize the parameters of the document models in order to significantly speed up the rate of convergence of the document models. 

\subsection{Data details}
Entities in the CoNLL-2003 corpus are labeled with one of four types: {\sc PER}, {\sc ORG}, {\sc LOC} or {\sc MISC}, with a fairly even distribution over the four entity types. OntoNotes contains a larger and more diverse set of 19 different entity types, adding: {\sc ORDINAL}, {\sc PRODUCT}, {\sc NORP}, {\sc WORK\_OF\_ART}, {\sc LANGUAGE}, {\sc MONEY}, {\sc PERCENT}, {\sc CARDINAL}, {\sc GPE}, {\sc TIME}, {\sc DATE}, {\sc FAC}, {\sc LAW}, {\sc EVENT} and {\sc QUANTITY}. The OntoNotes corpus also covers a wider range of text genres, including telephone conversations, web text, broadcast news and translated documents, whereas the CoNLL-2003 text covers only newswire. The combined entity types and boundary encodings result in 17 possible output labels in the CoNLL-2003 corpus and 74 labels in the OntoNotes corpus. The sizes of the two corpora in terms of documents, sentences, tokens and entities are given in Table \ref{data-table}. 

\begin{table}
\begin{tabular}{p{2.3cm}p{0.5cm}p{1.3cm}p{1cm}p{1cm}}
Data & & Train & Dev & Test \\ \hline \hline
CoNLL-2003 & 
Tok \newline Sent \newline Doc \newline Ent &
204,567 \newline 14,041 \newline 945 \newline 23,499 & 
51,578 \newline 3,250 \newline 215 \newline 5,942 &
46,666 \newline 3,453 \newline 230 \newline 5,648\\ \hline
OntoNotes 5.0 & 
Tok \newline Sent \newline Doc \newline Ent &
1,088,503 \newline 59,924 \newline 2,483 \newline 81,828 & 
147,724 \newline 8,528 \newline 319 \newline 11,066 & 
152,728 \newline 8,262 \newline 322 \newline 11,257 \\
\end{tabular}

\caption{Statistics of NER datasets \label{data-table}}
\end{table} 

\subsection{Evaluation}

To select hyperparameters, we iteratively perform grid search over increasingly fine-grained settings of dropout, learning rate, Adam $\beta_2$ and $\epsilon$ parameters, gradient clipping threshold, number of dilated layers, number of repeated blocks, regularizer penalty and batch size. Since we found the variance in score between runs to vary significantly, in the last iteration of grid search, we ran each setting of parameters three times and averaged their scores on the validation set. Of these, we ran the top ten settings ten times, and took the parameters which averaged the highest F1 on the development set, and report scores on the test set using these parameters. Note that we do not in the final stage include the development set as training data as has been done in some previous work, and so do not directly compare to results from other papers which do so.

We evaluate test-time speed using our top-performing trained models. All timing experiments were run on a nVidia Titan X GPU with a 2.4GHz Intel Xeon CPU. We do not include data loading, preprocessing or feature hashing in our timing since this is exactly the same across all models. Reported is the time it takes for each model to produce a sequence of labels given a sequence of integers representing the words and their capitalization. After a burn-in run to account for caching and GPU data I/O, we run each model 20 times over the development set and average these times. We do this for batch sizes ranging from 1 to 10,000. 

\end{document}